\documentclass[runningheads]{llncs}
\usepackage[T1]{fontenc}
\usepackage{graphicx}
\usepackage{booktabs}
\usepackage[misc]{ifsym}
\newcommand{\corr}{(\Letter)}
\usepackage{xcolor}

\usepackage{amssymb}

\usepackage{hyperref}
\usepackage{url}
\usepackage{multirow}
\usepackage{caption}
\usepackage{subcaption}
\usepackage{microtype}
\usepackage{amsmath}

\usepackage{mwe}
\newcommand{\name}{\textbf{\texttt{NM-Transformer}}}

\usepackage{authblk}
\usepackage{footmisc}

\begin{document}

\title{Boosting Protein Language Models with Negative Sample Mining}
\toctitle{Boosting Protein Language Models with Negative Sample Mining} 


\author{Yaoyao Xu\thanks{Yaoyao Xu and Xinjian Zhao contributed equally to this paper.} \and Xinjian Zhao$^*$ \and Xiaozhuang Song \and Benyou Wang \and Tianshu Yu\corr}

\authorrunning{Y. Xu et al.}

\institute{School of Data Science, The Chinese University of Hong Kong, Shenzhen\\
\email{\{xuyaoyao, wangbenyou, yutianshu\}@cuhk.edu.cn}
\\
\email{\{xinjianzhao1, xiaozhuangsong1\}@link.cuhk.edu.cn}
}

\tocauthor{Yaoyao Xu, Xinjian Zhao, Xiaozhuang Song, Benyou Wang, Tianshu Yu}

\maketitle              

\begin{abstract}
We introduce a pioneering methodology for boosting large language models in the domain of protein representation learning. Our primary contribution lies in the refinement process for correlating the over-reliance on co-evolution knowledge, in a way that networks are trained to distill invaluable insights from negative samples, constituted by protein pairs sourced from disparate categories. By capitalizing on this novel approach, our technique steers the training of transformer-based models within the attention score space. This advanced strategy not only amplifies performance but also reflects the nuanced biological behaviors exhibited by proteins, offering aligned evidence with traditional biological mechanisms such as protein-protein interaction. We experimentally observed improved performance on various tasks over datasets, on top of several well-established large protein models. This innovative paradigm opens up promising horizons for further progress in the realms of protein research and computational biology. The code is open-sourced at \url{https://github.com/LOGO-CUHKSZ/NM-Transformer}.

\end{abstract}

\section{Introduction}
The advancements in machine learning and artificial intelligence catalyze a transformative shift in natural science research, offering significant advantages in tackling critical scientific challenges~\cite{wang2023scientific}.
While traditional approaches often rely on costly experimental and computational procedures, data-driven learning paradigms provide a cost-effective alternative by leveraging the ability to extract specific patterns from historical data. These paradigms accelerate the application of insights to scientific discovery and minimize the need for extensive experimental processes.
%
Among various scientific domains benefiting from such advancements, the field of proteins stands out significantly. The continuous expansion of experimental protein structures, meticulously cataloged in the Protein Data Bank (PDB)~\cite{burley2017protein} — a cornerstone for applying data-driven methodologies — exemplifies the synergy between large-scale biological data and artificial intelligence. This convergence has propelled the protein field to the forefront of disciplines integrating AI technologies, that is, AI for Proteins (AI4Proteins). 

AI4proteins focuses on utilizing abundant protein data for representation learning, extracting important features from raw data presented in sequential (e.g., FASTA) or structured (graph) formats.~\cite{gligorijevic2021structure,strokach2020fast}.
Recently, transformer-based Large Language Models (LLMs) have become dominant in the representation learning of proteins by incorporating further knowledge from the biological domain, demonstrating significant promise. The acquired representations have been effectively employed in diverse downstream tasks, encompassing protein function prediction~\cite{gligorijevic2021structure,bonetta2020machine}, protein sequences design~\cite{ferruz2022controllable}, and prediction of protein folding structures~\cite{jumper2021highly,fang2023method}. 
To inject biological knowledge into LLMs, existing successful models generally converge to introducing ``co-evolution'' during the pre-training phase in either explicit or implicit fashion~\cite{meng2023improved,zhang2024protein}. The core of co-evolutionary methods lies in the interconnectedness of protein evolution, where the mutation of one amino acid can lead to correlated mutations in other amino acids to maintain the stability and function of the protein structure within three-dimensional space~\cite{jones2012psicov,de2013emerging,marks2011protein}. Therefore, sequence variations within protein families serve as informative indicators of protein 3D structure, and constraints on protein structure can also be inferred from patterns in homologous sequences. 

Explicitly incorporating co-evolution knowledge in LLMs is straightforward. For example, AlphaFold2~\cite{jumper2021highly} and Multiple Sequence Alignment (MSA) transformer~\cite{rao2021msa} operate by inputting aligned sequences from multiple sequences, containing evolutionary coupling data between amino acid residues within a structure.
On the other hand, ESMs~\cite{lin2023evolutionary} have shown promise by replacing the modeling approach of MSA to enable the direct prediction of protein folding structure and function using individual sequences. ESMs seemingly unravel the inherent dynamics of a single amino acid residue sequence and have learned protein folding from physics without any co-evolution signals. However, recent research~\cite{zhang2024protein} provides ample evidence that, although ESMs do not overtly rely on co-evolutionary signals, they retain co-evolutionary information through an implicit mechanism. Their analysis of the categorical Jacobian reveals that ESM-2~\cite{lin2022language} stores statistics of co-evolving residues similar to simpler models like Markov Random Fields and Multivariate Gaussian models. Additionally, the study identifies instances where ESM-2 incorrectly utilizes co-evolutionary information to predict the folded structure of an isomer sequence within the full-length context of a protein, strongly reinforcing this observation.



Thus, it is evident that the excellent performance of protein language models (PLMs) on structure-oriented tasks can be attributed to the outstanding understanding of co-evolutionary knowledge. However, aside from structures, co-evolutionary signals cannot clearly capture protein function and other important characteristics~\cite{xu2023protst}. For example, in many downstream tasks that are not directly related to co-evolutionary signals, such as subcellular localization, directly applying these PLMs results in suboptimal performance. Moreover, PLMs tend to be overly reliant on homologous sequences for inference due to their sensitivity to co-evolutionary signals, neglecting the comprehension of pattern distinctions among non-homologous sequences~\cite{bepler2021learning}.
Therefore, it is desirable to design routines to correct the overexposure of PLMs to co-evolutionary knowledge.

To achieve this, in this paper we resort to negative mining during the fine-tuning phase to alleviate the bias towards co-evolution in pre-trained PLMs. While in most contexts negative mining refers to enlarging the disparity with different labels in the \emph{feature space}~\cite{ying2018graph}, the proposed negative mining instead innovatively manipulates to decrease the alignment in the \emph{attention space} by enforcing the cross query-key response matrix to be uniform for negative pair, we name this framework as \textbf{\texttt{NM-Transformer}}. This design is built on the assumption that proteins with different labels orthogonal to co-evolution should not be well aligned by any means. Empowered by negative-mining-based fine-tuning, several PLMs empirically demonstrated substantial improvement on various downstream tasks compared to naive fine-tuning. Notably, in the protein-protein interaction task, we found strong evidence that negative mining enhances the alignment of amino acid residues at the binding boundary. This in turn supports our assumption with much interpretability.
In conclusion, our contributions are: 
      \textbf{(1)}. We propose a novel alignment-based negative mining framework for fine-tuning PLMs, which alleviates the issue where PLMs are overly reliant on co-evolution and struggle to swiftly transfer to downstream tasks.
     \textbf{(2)}. Our method is highly interpretable, enabling the identification of key residue sites near the binding surface that play a significant role in Protein-Protein Interactions (PPIs).
     \textbf{(3)}. Our experiments demonstrate that incorporating negative sample mining during the fine-tuning phase notably enhances the performance of PLMs in protein-pair and protein-wise tasks. Furthermore, our negative sample mining approach can bridge the performance gap between small-scale and large-scale PLMs, highlighting the potential applications of our methodology in resource-constrained environments.

\section{Related Work}
\noindent \textbf{Protein Language Model.} Recent advancements in natural language processing have introduced large Transformer models, such as BERT~\cite{devlin2018bert}, which have gained widespread adoption for protein-related tasks~\cite{rives2021biological,rao2019evaluating,elnaggar2021prottrans}. 
The use of PLMs, trained on extensive sequence databases, has shown success in various protein-related endeavors, including secondary structure prediction~\cite{rao2019evaluating,heinzinger2019modeling,rives2021biological,elnaggar2021prottrans}, contact prediction~\cite{rao2020transformer,elnaggar2021prottrans}, 3D structure prediction~\cite{jumper2021highly,baek2021accurate},protein-protein interaction prediction~\cite{evans2021protein}, and fitness prediction~\cite{alley2019unified,dallago2021flip,hie2021learning,hie2022evolutionary,baek2021accurate}.
Existing models for protein PLMs can be categorized based on their attention mechanisms: evolution-aware and evolution-free PLMs~\cite{hu2022exploring}. Evolution-aware PLMs, like MSA-Transformer~\cite{rao2021msa} and Evoformer~\cite{jumper2021highly}, are designed to handle evolution-related protein sequences aligned through multiple sequence alignments (MSA).  
In contrast, evolution-free PLMs, such as ESM-1b, ESM-2~\cite{lin2023evolutionary}, and TAPE~\cite{rao2019evaluating}, operate on individual protein sequences without considering their evolutionary relationships. Given the protein sequences’ primary structure, their 3D structure could be generated through inverse folding~\cite{hsu2022learning}, and proteins’ functional properties are largely determined by their 3D structure. PLMs play a crucial role in encoding information from the primary and secondary structures of proteins. Notably, ESM-1b has demonstrated SOTA performance in predicting protein structure and functions. Building upon ESM-1b, ESM-2 incorporates improvements in model architecture, training parameters, computational resources, and data. ESM-2 represents one of the most advanced PLMs up to date. Considerable studies in protein representation learning employ ESM-2 as an encoder to acquire the primary structure representation of proteins. For instance, LM-Design~\cite{zheng2023structure}  utilized an additional structure encoder to augment ESM-2 with an understanding of protein structure, enabling the design of protein sequences with desired folds. ProtST~\cite{xu2023protst} involves fine-tuning ESM-2 using biomedical texts to learn protein representations while GearNet~\cite{zhang2022protein} pre-train protein graph encoder with the protein representations encodes from ESM-2. In this work, we focus on fine-tuning various scaled ESM-2 models for protein-wise classification and protein-protein interaction predictions. We employ a negative sample mining method to enhance the learning of protein representations. By leveraging these techniques, we aim to achieve improved performance in these tasks.\\

\noindent \textbf{Negative Samples Mining.} Negative samples mining is a commonly employed technique in representation learning, demonstrating effectiveness across a variety of real-world applications including object detection~\cite{shrivastava2016training}, answering systems~\cite{rao2016noise}, and recommender systems~\cite{wang2020reinforced}. By improving the model's discrimination between a target sample and its negative counterpart, negative sampling enhances its ability to discern more precise decision boundaries. In the field of AI4Proteins, negative sample mining techniques have also been explored and validated for their effectiveness. For instance, CLEAN~\cite{yu2023enzyme} significantly enhanced enzyme function prediction by employing contrastive learning to fine-tune the ESM encoder, with negative samples defined based on enzyme commission numbers (EC numbers)~\cite{bairoch2000enzyme}. Some research has also considered sampling a large number of random negative samples for protein representation learning~\cite{wei2017improved,wang2024hierarchical}, however, due to the emergence of powerful protein language models, the use of a small number of explicitly labeled negative samples for further representation learning on top of PLMs has become a more promising direction.

\begin{figure}[t!]
    \centering
    \resizebox{1\textwidth}{!}{
    \includegraphics{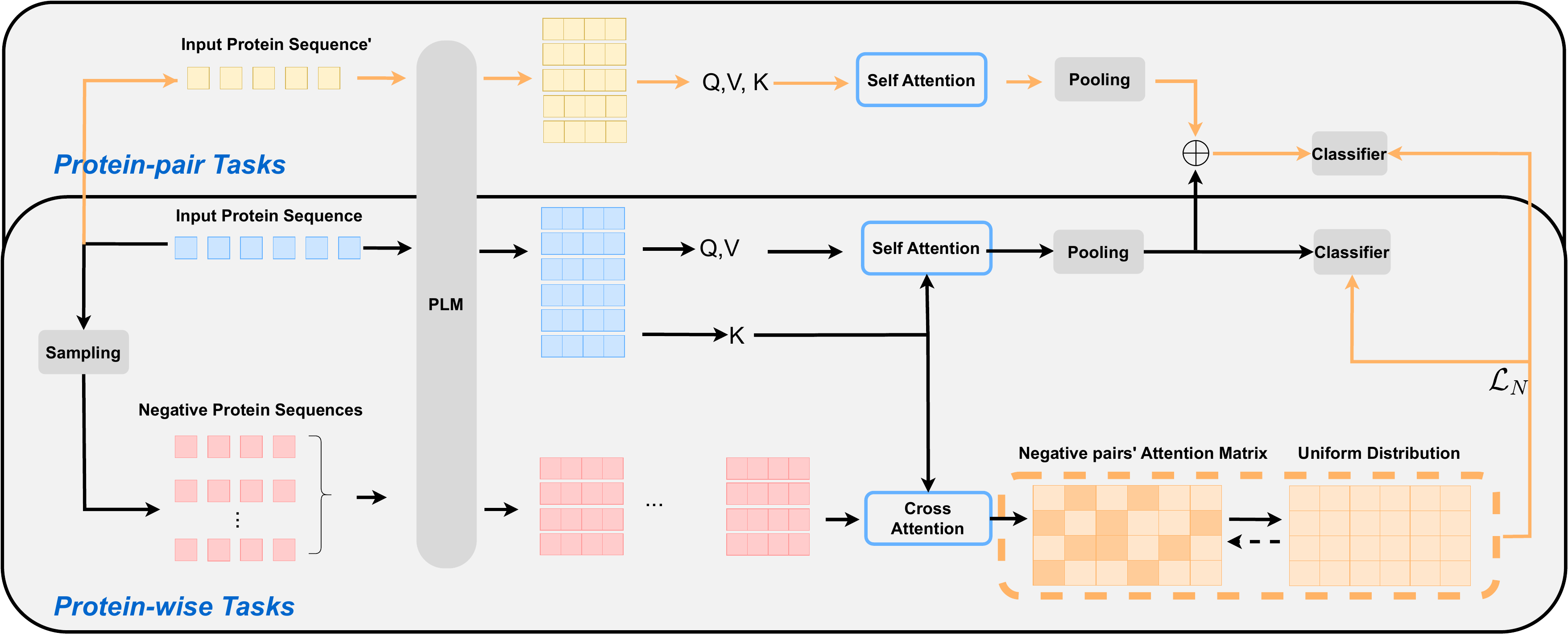}}
    \caption{~\name~ framework. Our framework is designed for protein-wise and protein-pair tasks, taking single protein sequences or pairs of protein sequences as inputs. It consists of two main steps: 1) Negative sampling: for protein-wise tasks, we sample proteins with differing properties as negative examples based on task-predicted properties (e.g., solubility); for protein-pair tasks, we sample non-interacting proteins as negative examples. 2) Negative sample mining: we optimize the cross-attention matrix to align with uniform distributions of input and negative sample sequences, guide PLMs to learn discriminative embeddings, and generate representations for downstream tasks using self-attention layers.}
    \label{fig:enter-label}
\end{figure}

\section{Method}
In this section, we introduce our framework \textbf{\texttt{NM-Transformer}} by negative sample mining that encourages PLMs to avoid aligning protein sequences that lack relevance within the attention space for a specific downstream task. By employing this technique, we alleviate the issue of PLMs relying on homologous protein sequences for inference under the influence of co-evolutionary signals, while neglecting the understanding of patterns in unrelated protein sequences~\cite{zhang2024protein,bepler2021learning}. The overall framework is shown in Figure~\ref{fig:enter-label}. In Subsection~\ref{negative}, Negative samples of proteins are defined for both the protein-pair and the protein-wise tasks, the representations of the protein sequences and their corresponding negative sample sequences are obtained using a PLM encoder. Subsequently, we elaborate on our proposed framework in detail in subsection~\ref{Negative Mining in Attention Space}, where our work forces protein sequences and their negative sample sequences to be unaligned in the attention space, ultimately producing a nearly uniform cross-attention matrix. At the same time, our work analyzes how the representation learned under this constraint can be helpful for protein-related tasks.

\subsection{Negative Sampling}\label{negative}
In this section, we introduce the negative sampling strategy for protein sequences, where each token of a protein sequence represents an amino acid residue.
For protein-wise tasks,  
given a specific protein sequence $\mathbf{s}_{g}$, with maximum length $m$. $N$ negative samples $\mathbf{s}_{n}$ are randomly sampled from the different labels from the input sequence with maximum length $l$.  For protein-pairwise tasks, negative samples are defined as a set of $N$ protein sequences that do not interact with the given $\mathbf{s}_{g}$. The sampling process can be represented as follows:

\begin{equation}
\begin{aligned}
\mathbf{s}_{n_i} & \sim D, & \text{where } \mathrm{label}(\mathbf{s}_{n_i}) \neq \mathrm{label}(\mathbf{s}_{g}) \quad  \text{(for protein-wise task)} \\
\mathbf{s}_{n_i} & \sim D, & \text{where } \mathrm{label}(\mathbf{s}_{n_i}, \mathbf{s}_{g}) \neq 1 \quad  \text{(for protein-pairwise task)}
\end{aligned}
\end{equation}
where $\mathbf{s}_{n_i}$ represents the $i$-th negative sample sampled from dataset $D$. In the protein-wise task, the $\mathrm{label}$ function identifies the protein type to which it belongs, while in the protein-pair task, $\mathrm{label}$ indicates the interaction status between two proteins: $1$ denotes interaction, and $0$ signifies no interaction.

The input sequence with their negative samples are fed into the PLM to obtain their corresponding embeddings, which is denoted as $\mathbf{E}_{g}\in\mathbb{R}^{m \times d}$ and $\mathbf{E}_{n}\in\mathbb{R}^{l \times d}$, as modeled in Equation~\ref{eq: Emb}. Here, $f_{\theta}$ denotes the PLMs encoder with parameters $\theta$, and $d$ is the hidden dimension of embedding.
\begin{equation}
\mathbf{E}_{g} = f_{\theta}\left(\mathbf{s}_{g}\right)
, \quad \mathbf{E}_{n} = f_{\theta}\left(\mathbf{s}_{n}\right)
\label{eq: Emb}
\end{equation}


\subsection{Negative Mining in Cross Attention Space}\label{Negative Mining in Attention Space} After obtaining the embedding of the PLM encoder, we hope that by imposing constraints between protein sequence representations, the PLM can be migrated more efficiently to downstream tasks during fine-tuning. Rather than directly manipulating features, our work instead proposes to reduce the alignment between protein sequences and their negative samples within the cross-attention space. This strategy is designed to guide PLMs in accurately discriminating between protein sequences and their negative counterparts within the representation space via alignment. Subsequently, the resulting attention scores can provide insights for comprehensive analysis and interpretation of protein tasks.

Concretely, the cross-attention matrix is constructed between the protein sequence and its corresponding negative sample sequence as follows:

\begin{equation}
\mathbf{K}_{g}=\mathbf{W}_{g}^{K}\cdot \mathbf{E}_{g}
, \quad \mathbf{Q}_{n}=\mathbf{W}_{n}^{Q}\cdot \mathbf{E}_{n} ,\quad \mathbf{V}_{g}=\mathbf{W}_{g}^{V}\cdot \mathbf{E}_{g}
\label{eq: attention0}
\end{equation}
\vspace{-1em}
\begin{equation}
\mathbf{Att}_{neg} =\operatorname{softmax}\left(\mathbf{Q}_{n}\mathbf{K}_{g}^{T}\right)
\label{eq: attention}
\end{equation}
Where the cross-attention matrix between the negative pairs, denoted as $\mathbf{Att}_{neg} \in \mathbb{R}^{m \times l}$, is computed using the key and value matrix $\mathbf{K}_{g}$ and $\mathbf{V}_{g}$ from the $\mathbf{E}_{g}$, query matrix $\mathbf{Q}_{n}$ from $\mathbf{E}_{n}$. After the softmax function, we maximize the likelihood between all negative cross-attention matrices and the uniform distribution to reduce the alignment of protein residues in negative pairs. 

The negative sample loss maximizes the likelihood between all negative pairs' cross-attention matrices $\mathbf{Att}_{neg}$ and the uniform distribution, which is defined in Equation~\ref{eg: negative loss}, where $N$ represents the number of negative samples for each input sequence. The final training objective comes from the summation of the supervised loss and the negative sample loss $\mathcal{L}_{N}$.
\begin{equation}
\mathcal{L}_{N} = \sum_{i=1}^{N}{\mathrm{MSE}\left(\left(\mathbf{Att}_{neg}^{i}\right),  \mathbf{U}\right)}
\label{eg: negative loss}
\end{equation}
Where $\mathbf{U} \in \mathbb{R}^{m \times l} $ is a uniform matrix.

We apply a self-attention layer to the embedding $\mathbf{E}_{g}$ obtained from PLM. The key vector used for self-attention shares the same parameters as the key vector used in cross-attention. This design choice enables us to optimize both self-attention and cross-attention matrices within the same latent space. After the self-attention function, mean pooling is applied to the hidden representation $\mathbf{h}_{g} \in \mathbb{R}^{m \times d}$ to get sequence representation $\mathbf{h}^{'}_{g} \in \mathbb{R}^{d}$. The supervised loss $\mathcal{L}_{S}$ calculates cross entropy between the protein's ground truth label and the project-to-class states after the MLP classifier as follows:

\begin{equation}
\mathbf{h}^{'}_{g} = \operatorname{Pooling} \left(\operatorname{softmax}\left(\frac{\mathbf{Q}_{g}\mathbf{K}_{g}^{T}}{\sqrt{d_{k}}}\right)\mathbf{V}_{g}\right), \quad \mathbf{h}^{'}_{g} \in \mathbb{R}^{d}
\end{equation}
\begin{equation}
\mathcal{L}_{S} = \operatorname{CE} \left( \operatorname{softmax} \left(\mathbf{W}\cdot \mathbf{h}^{'}_{g} + \mathbf{b}\right),  \quad  y_{g} \right)
\label{eg: supervised loss}
\end{equation}

The total loss $\mathcal{L}_{total}$ of our framework can be represented as follows:

\begin{equation}
\mathcal{L}_{total} = \mathcal{L}_{S} + \mathcal{L}_{N}
\end{equation}
For protein-pair tasks that require paired inputs $\left(\mathbf{h}_{g}^{A}, \mathbf{h}_{g}^{B}\right)$, the training process is roughly the same as the above framework. The subtle differences are: (1) The negative pairs are the protein pairs that do not interact with each other and positive pairs interact with each other according to their ground truth label in each batch. (2) The self-attention will apply to a pair proteins and the concatenation is employed to the paired protein sequences' hidden states after the pooling function:
\begin{equation}
    \mathbf{h}_{g}^{'} = \mathbf{h}_{g}^{A} \oplus \mathbf{h}_{g}^{B}, \quad \mathbf{h}_{g}^{'} \in \mathbb{R}^{2\cdot d}
\end{equation}
Where $A$ and $B$ denote the paired two protein sequences. For the protein-pair task, the calculation of the loss is consistent with the protein-wise task.\\
\subsection{Inference Phase of NM-Transformer} 
The inference stage of our methodology presents a lightweight approach distinct from the training phase. Specifically, it omits the employment of sampling and cross-attention mechanisms, instead on a direct application of the PLMs and self-attention layer followed by a classifier. The weights of the self-attention layer and PLMs encoder are updated by negative sample mining and supervised loss, enabling the migration of protein representation learning from co-evolutionary predominance to adaptation to specific downstream tasks. 

\section{Experiments}
In this section, we conduct expensive experiments to evaluate the effectiveness of~\name~on pre-trained protein language models. Subsection~\ref{setting} introduces the experimental setup. Subsection~\ref{main result} showcases the performance results of both~\name~and the baseline methods. Furthermore, in Subsection~\ref{case study}, our analysis demonstrates that~\name~can generate a more reasonable cross-attention matrix compared to a basic transformer. Additionally, we delve into the interpretation of the attention score of~\name~through an in-depth analysis of two protein complexes as case studies.

\subsection{Experimental Settings}\label{setting}
\begin{table}[tp!]
    \centering \caption{Statistics of the datasets, including the number of sequences in the training, validation, and test sets, as well as the number of classes in each dataset.}
        \begin{tabular}{lccc}
            \toprule
            \textbf{Dataset} & \textbf{Seq. (Train/Val/Test)} & \textbf{Classes} \\
            \midrule
            Sol & 62478 / 6942 / 300 & 2 \\
            Sub  & 8420 / 2811 / 2773 & 10 \\
            Fold  & 12312 / 736 / 718 & 1195 \\
            Human  & 35669 / 315 / 237 & 2 \\
            Yeast  & 4945 / 95 / 394 & 2 \\
            \bottomrule
        \end{tabular}
    \label{dataset_statistics}
\end{table}

\noindent \textbf{Datasets.} Our work utilizes five datasets from the PEER benchmark~\cite{xu2022peer} to assess the effectiveness of NM (Negative Sample Mining). Specifically, our experiments are conducted on two Protein-Protein Interaction (PPI) prediction datasets: Human PPI and Yeast PPI as well as three Protein-wise classification datasets: Subcellular localization, Fold, and Solubility. Here, we briefly describe the tasks corresponding to the five datasets. \underline{\textit{(1) Subcellular localization: }}Involves predicting the subcellular localization of natural proteins within a cell. Accurate determination of a protein's subcellular localization significantly enhances target identification in the context of drug discovery~\cite{rajagopal2003subcellular}. Each protein is assigned a categorical label, such as ``lysosome'', indicating its specific location. There are 10 possible localizations, denoted as labels $y \in {0,1, ..., 9}$. \underline{\textit{(2) Fold classification: }}Involves categorizing the overall structural topology of a protein at the fold level. This categorization is represented by a categorical label, denoted as $y$, and it can take values from 0 to 1194. Folding is important for both functional analysis and drug design~\cite{chen2016profold}, and their label is determined based on the backbone coordinates of the protein structure. \underline{\textit{(3) Solubility Prediction: }}Forecast the solubility of a protein, determining whether it falls into the category of soluble or not (labeled as $y \in \left(0,1\right)$). The solubility of proteins is important in the domain of pharmaceutical research and industry, as it is a vital characteristic for the effectiveness of functional proteins~\cite{khurana2018deepsol}. \underline{\textit{(4) Yeast PPI: }}Predicts whether two yeast proteins interact or not. The negative pairs are from different subcellular locations. \underline{\textit{(5) Human PPI: }}Predicts whether two human proteins interact or not. Negative pairs are from different subcellular locations. 
All datasets employ accuracy as the evaluation metric following the PEER benchmark. For detailed statistical information regarding the datasets, kindly refer to Table~\ref{dataset_statistics}.\\

\noindent \textbf{Baselines.} 
We compare our method with baselines spanning protein sequence encoders and pretrained protein language models:
\begin{itemize}
    \item \textbf{Protein Sequence Encoders ~\cite{rao2019evaluating}:}
    \begin{itemize}
        \item \textbf{LSTM:} Utilizes a recurrent neural network architecture to capture long-range interactions within protein sequences.
        \item \textbf{Transformer:} Employs a self-attention mechanism to model both short-range and long-range dependencies in sequences.
        \item \textbf{ResNet:} Designed for capturing short-range interactions through deep convolutional network architecture.
    \end{itemize}
    \item \textbf{Pretrained Protein Language Models (PLMs):}
    \begin{itemize}
        \item \textbf{ProtBert~\cite{ProtBert}:} A Transformer-based model pre-trained on over 2.1 billion protein sequences, designed to understand complex protein language patterns using the masked language modeling technique.
        \item \textbf{ESM-2~\cite{lin2022language}:} The most advanced Transformer-based model pre-trained on over 24 million protein sequences, utilizing the same MLM technique for capturing the essence of protein sequences.
    \end{itemize}
\end{itemize}
We assess the performance of \texttt{NM-Transformer} using both fixed PLMs and fine-tuned PLMs in each task. Only the best performances are reported for the sake of brevity. 
In \texttt{NM-Transformer}, we introduce a self-attention layer i.e., the Transformer layer, after the PLM to obtain the final representation for the downstream task(refer to Section~\ref{Negative Mining in Attention Space}). To demonstrate the effectiveness of our approach, which focuses on mining negative sample information instead of incorporating a self-attention mechanism after the PLM, ablation experiments are conducted to compare the performance of our method with the following approaches: 1). Transformer Classifier: Includes a self-attention layer after the PLM, but it does not compute cross-attention between negative pairs and only optimizes the supervised loss. 2). MLP Classifier: Utilizes an MLP layer as a classifier and solely optimizes the supervised loss, following the traditional fine-tuning method. By comparing our method with these baselines, we observed significant improvements achieved by our \texttt{NM-Transformer} in all five tasks.

\noindent \textbf{Implementation.} The embeddings for protein sequences are obtained using ESM-2 with 8M, 35M, and 150M number of parameters and ProtBret with 420M number of parameters. The maximum protein sequence length is $550$, with a hidden dimension of the projected protein embedding and classifier to be $128$. 
All experiments are conducted on 8 NVIDIA A100 GPUs (40GB).

\subsection{Main results}\label{main result}
We compare the PPI prediction accuracy in Table~\ref{result_ppi} and protein-wise classification accuracy performance of different methods on three datasets in Table~\ref{result}. The results show that ~\name~achieves consistent performance gains across all datasets and PLMs with different scales, demonstrating the effectiveness of our approach. 
It is noteworthy that utilizing the ~\name~ or employing the transformer as a downstream classifier enables smaller PLMs to approach the performance levels of larger PLMs. 

\begin{figure}[p!]
    \centering
    \resizebox{0.85\textwidth}{!}{
    \includegraphics{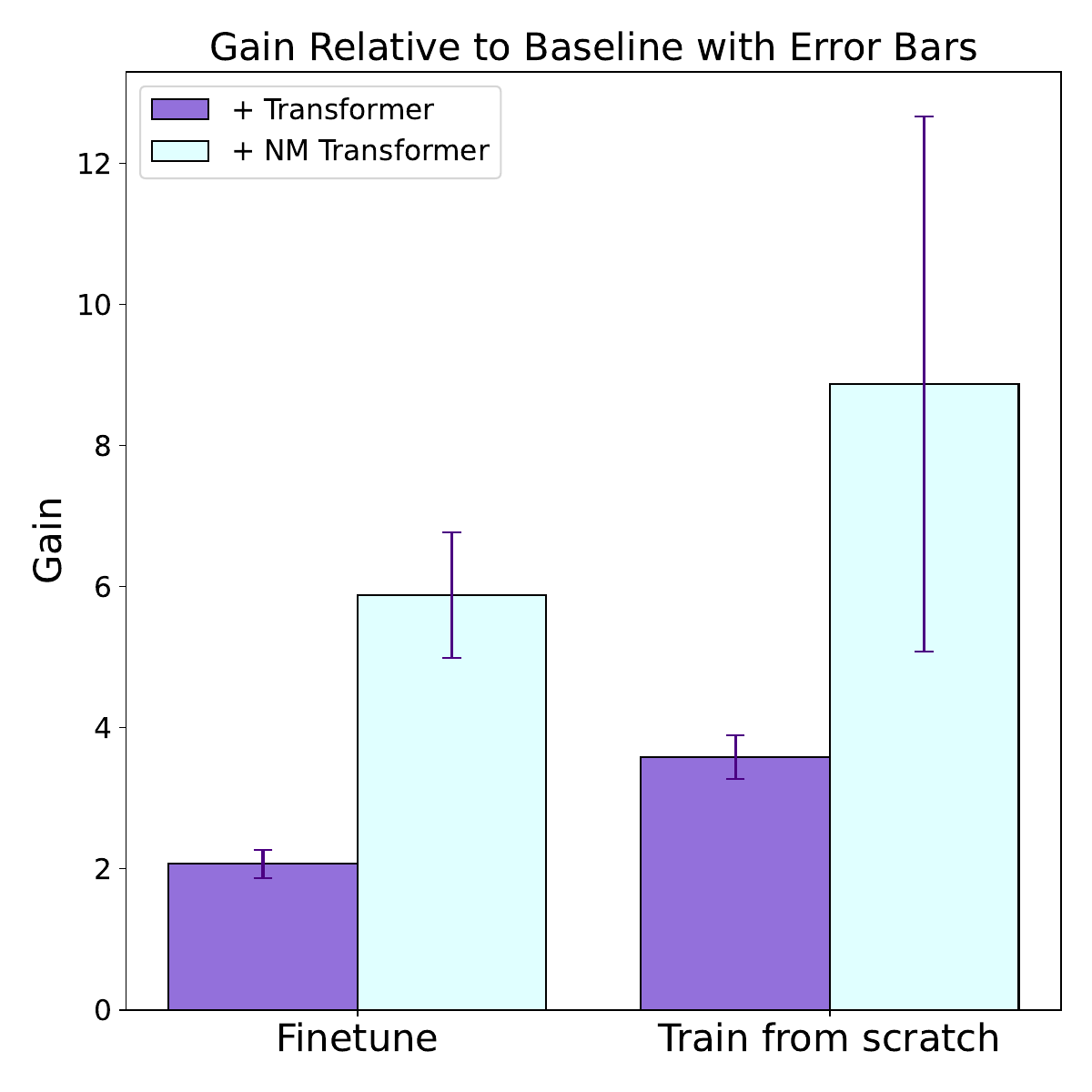}
    \includegraphics{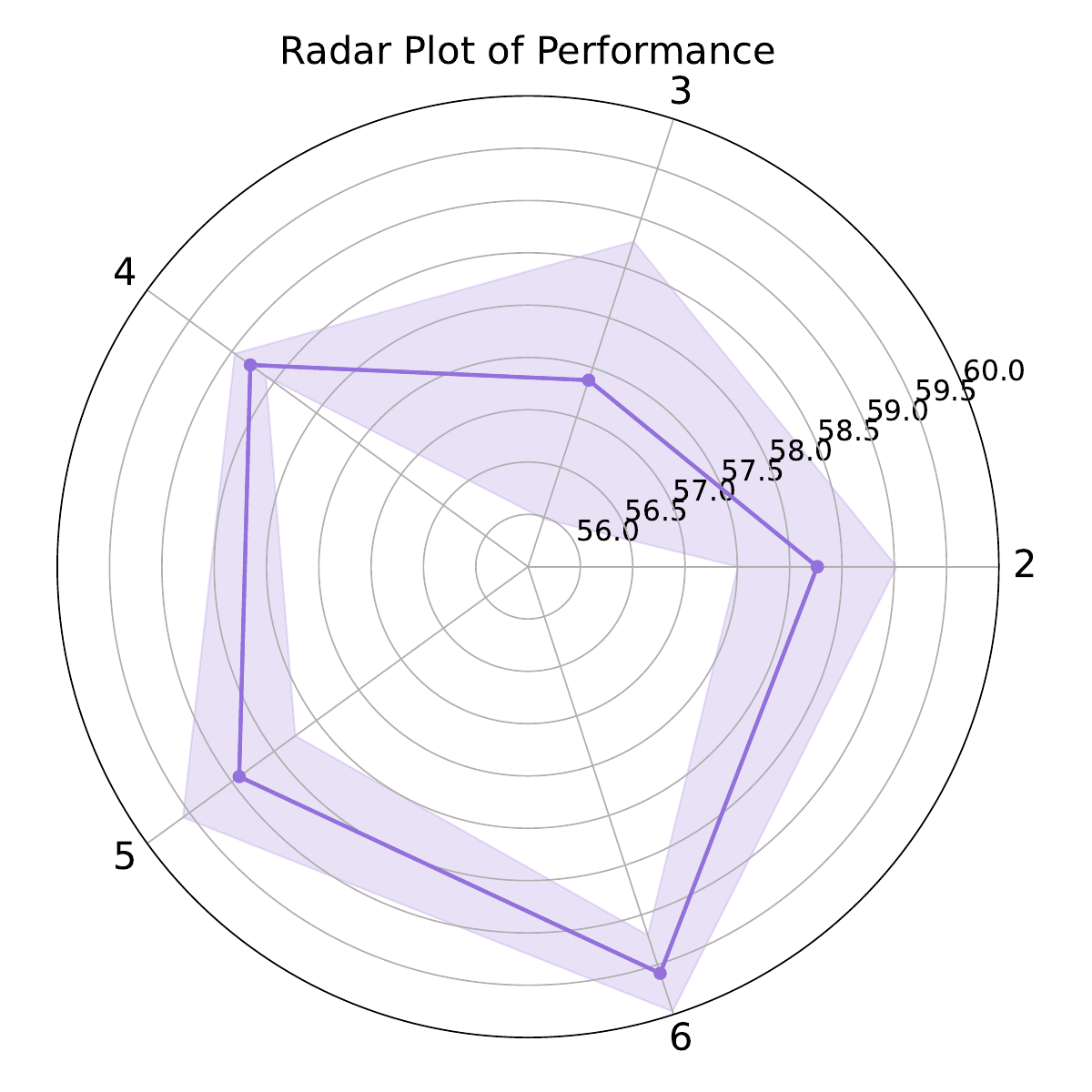}}
    \caption{The histogram on the left demonstrates the performance improvement relative to MLP when training from scratch and fine-tuning using~\name~and Transformer. The right radar figure illustrates the performance corresponding to the number of negative samples, as the number of negative samples increases, the performance of~\name~continues to improve. The experiments were all run on the Sub dataset using ESM-2(8M).}
    \label{fig: pretrain}\vspace{-3mm}
\end{figure}

Moreover, the differential impact of our method on well-trained PLMs compared to randomly initialized PLMs is examined. Through experiments on the subcellular localization dataset, utilizing ESM-2(8M) as a case study in Figure~\ref{fig: pretrain}, we observed significant enhancements in the performance of both finetuning PLMs and training PLMs from scratch, particularly in the case of training the PLMs from scratch. Based on our observations, it could be concluded that the co-evolutionary information acquired by the PLM during the pre-training phase enhances protein representation. Additionally, the incorporation of negative sample mining effectively rectifies the deficiencies introduced by the co-evolutionary information, particularly in downstream tasks like protein function prediction. Consequently, when training PLM from scratch for these tasks, although our approach yields a greater performance improvement compared to fine-tuning PLM, the ultimate performance falls short of that achieved by incorporating negative mining to rectify co-evolutionary information in PLM.
Additionally, we studied the influence of the number of negative samples per protein sequence on performance by evaluating performance variations with different numbers of negative samples in the subcellular localization dataset, depicted in Figure~\ref{fig: pretrain} using radar plots. Our results suggest a positive correlation between the augmentation of negative samples and the model's effectiveness.

\begin{table*}[tp!]
    \centering   \caption{Performance comparison of various models in protein-protein interaction predictions: Yeast PPI, and Human PPI.  $\left[\dag\right]$ denotes results taken from PEER~\cite{xu2022peer}.}
    \begin{sc}
        \begin{tabular}{cccccc}
            \toprule
            \textbf{Model} & \textbf{Classifier} & \textbf{Human PPI} & \textbf{Yeast PPI} \\
            \midrule
            \multicolumn{4}{c}{\textbf{Protein Sequence Encoders}} \\
            \midrule
            LSTM$^{\dag}$ & MLP & 63.75 $\pm$ 5.12 & 53.62 $\pm$ 2.72 \\
            Transformer$^{\dag}$ & MLP & 59.58 $\pm$ 2.09 & 54.12 $\pm$ 1.27\\
            ResNet$^{\dag}$ & MLP & 68.61 $\pm$ 3.78 & 48.91 $\pm$ 1.78 \\
            \midrule
            \multicolumn{4}{c}{\textbf{PLM}} \\
            \midrule
            \multirow{3}{*}{ESM2 8M} & MLP &  79.60 $\pm$ 0.19 & 59.47 $\pm$ 0.43  \\
            & Transformer & 79.04 $\pm$ 0.39 & 59.89 $\pm$ 1.90  \\
            & \name & \textbf{81.57 $\pm$ 0.86} & \textbf{60.65 $\pm$ 1.45}  \\
            \midrule
            \multirow{3}{*}{ESM2 35M} & MLP & 86.07 $\pm$ 2.01  & 57.95 $\pm$ 0.93 \\
            & Transformer & 85.86 $\pm$ 1.47& 58.03 $\pm$ 2.01 \\
            & \name &  \textbf{87.34 $\pm$ 0.42} &  \textbf{63.19 $\pm$ 0.90} \\
            \midrule
            \multirow{3}{*}{ESM2 150M} & MLP &  87.76 $\pm$ 0.42& 58.46 $\pm$ 0.31  \\
            & Transformer & 87.34 $\pm$ 0.84 & 62.60 $\pm$ 0.93 \\
            & \name &  \textbf{88.39 $\pm$ 0.21} &  \textbf{65.31 $\pm$ 0.83} \\
            \midrule
            \multirow{1}{*}{ESM-1b 650M$^{\dag}$} & MLP &  88.06 $\pm$ 0.24 & 66.07 $\pm$ 0.58 \\
            \midrule
            \multirow{3}{*}{ProtBert} & MLP & 84.38 $\pm$ 0.42 &  60.64 $\pm$ 1.14\\
            & Transformer & 84.81 $\pm$ 0.42 & 61.16 $\pm$ 0.64 \\
            & \name & \textbf{85.44 $\pm$ 1.05}  &  \textbf{62.18 $\pm$ 0.57} \\
            \bottomrule
        \end{tabular}
    \end{sc}
    \label{result_ppi}
\end{table*}

\begin{table*}[tp!]
    \centering  \caption{Performance comparison of various models in protein-wise tasks. Results are reported in terms of accuracy and standard deviation.  $\left[\dag\right]$ denotes results taken from PEER~\cite{xu2022peer}.}
    \begin{sc}
        \begin{tabular}{cccccc}
            \toprule
            \textbf{Model} & \textbf{Classifier} & \textbf{Sol (2)} & \textbf{Sub (10)} & \textbf{Fold (1195)} \\
            \midrule
            \multicolumn{5}{c}{\textbf{Protein Sequence Encoders}} \\
            \midrule
            LSTM$^{\dag}$ & MLP & 70.18 $\pm$ 0.63 &  62.98 $\pm$ 0.37 & 8.24 $\pm$ 1.61 \\
            Transformer$^{\dag}$ & MLP & 70.12 $\pm$ 0.31 & 56.02 $\pm$ 0.82 & 8.52 $\pm$ 0.63 \\
            ResNet$^{\dag}$ & MLP &  67.33 $\pm$ 1.46 & 52.30 $\pm$ 3.51 & 8.89 $\pm$ 1.45 \\
            \midrule
            \multicolumn{5}{c}{\textbf{PLM}} \\
            \midrule
            \multirow{3}{*}{ESM2 8M} & MLP & 64.23 $\pm$ 0.40 & 68.66 $\pm$ 0.38 & 22.56 $\pm$ 0.79\\
            & Transformer & 73.28 $\pm$ 2.02 & 71.06 $\pm$ 0.61 & 22.23 $\pm$ 0.73 \\
            & \name & \textbf{73.84 $\pm$ 1.28} & \textbf{72.24 $\pm$ 0.40} & \textbf{23.12 $\pm$ 0.59} \\
            \midrule
            \multirow{3}{*}{ESM2 35M} & MLP & 62.18 $\pm$ 1.00 & 72.88 $\pm$ 1.01 & 25.20 $\pm$ 0.39 \\
            & Transformer & 73.50 $\pm$ 1.40 & 73.47 $\pm$ 0.90 & 27.21 $\pm$ 0.65 \\
            & \name & \textbf{74.18 $\pm$ 0.86} & \textbf{75.19 $\pm$ 0.64} & \textbf{27.72 $\pm$ 0.52} \\
            \midrule
            \multirow{3}{*}{ESM2 150M} & MLP & 65.84 $\pm$ 1.27 & 75.73 $\pm$ 0.56 & 26.27 $\pm$ 1.03 \\
            & Transformer & 74.67 $\pm$ 0.74 & 76.45 $\pm$ 1.78 & 26.60 $\pm$ 1.49 \\
            & \name & \textbf{75.16 $\pm$ 0.25} & \textbf{76.90 $\pm$ 1.36} & \textbf{27.11 $\pm$ 1.31} \\
            \midrule
            \multirow{1}{*}{ESM-1b 650M$^{\dag}$} & MLP & 70.23 $\pm$ 0.75 & 78.13 $\pm$ 0.49 & 28.17 $\pm$ 2.05 \\
            \midrule
            \multirow{3}{*}{ProtBert} & MLP &68.02 $\pm$ 0.35 &74.75 $\pm$ 0.80 & 18.98 $\pm$ 1.08 \\
            & Transformer &  73.40 $\pm$ 0.61& 74.04 $\pm$ 0.42& 20.89 $\pm$ 1.50 \\
            & \name & \textbf{73.93 $\pm$ 0.78} & \textbf{74.97 $\pm$ 0.56}& \textbf{21.54 $\pm$ 0.57} \\
            \bottomrule
        \end{tabular}
    \end{sc}
    \label{result}
\end{table*}

\subsection{Interpretability of~\name, Case Study}
\label{case study} 
\begin{figure}[pt!]
    \centering\resizebox{0.98\textwidth}{!}{
    \includegraphics{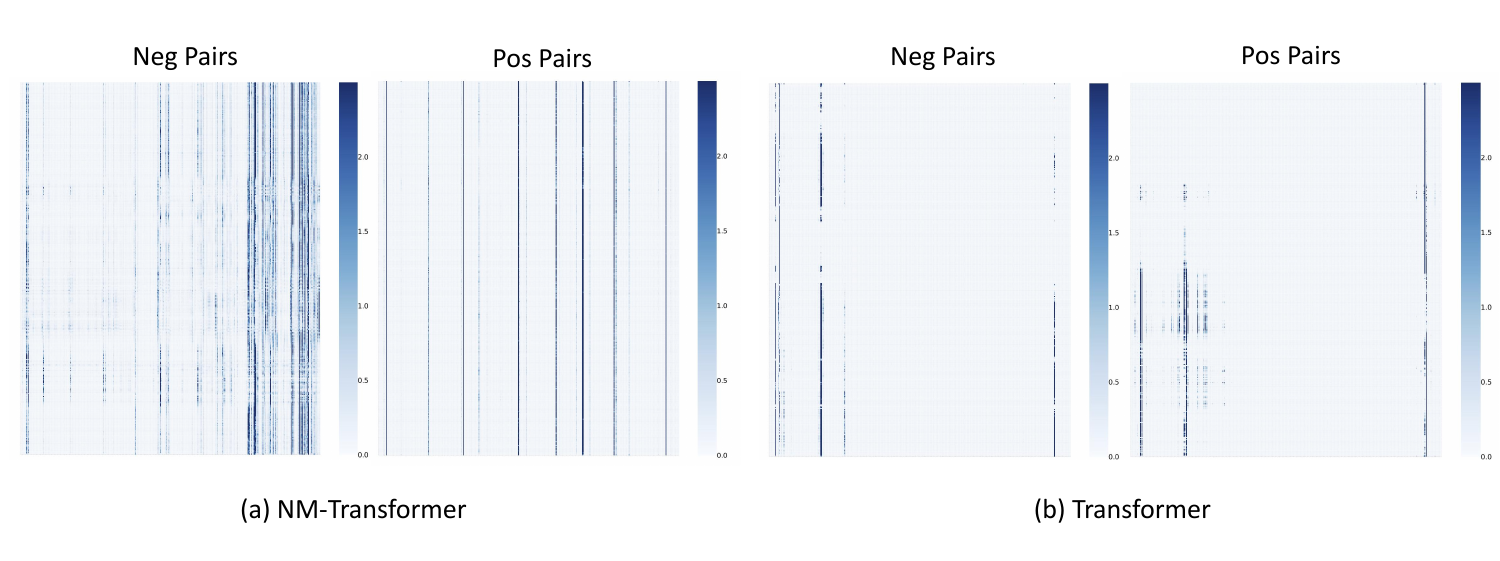}}
    \caption{Subfigures (a) and (b) show the cross-attention matrices produced by \name~and Transformer, highlighting clear differences between positive and negative pairs in our approach's matrix that are absent in the Transformer model's matrix. The regions surpassing the average attention score threshold are marked in the deepest shade of blue. }
    \label{fig: attention matrix}
\end{figure}

\begin{figure}[t!]
    \centering\resizebox{0.95\textwidth}{!}{
    \includegraphics{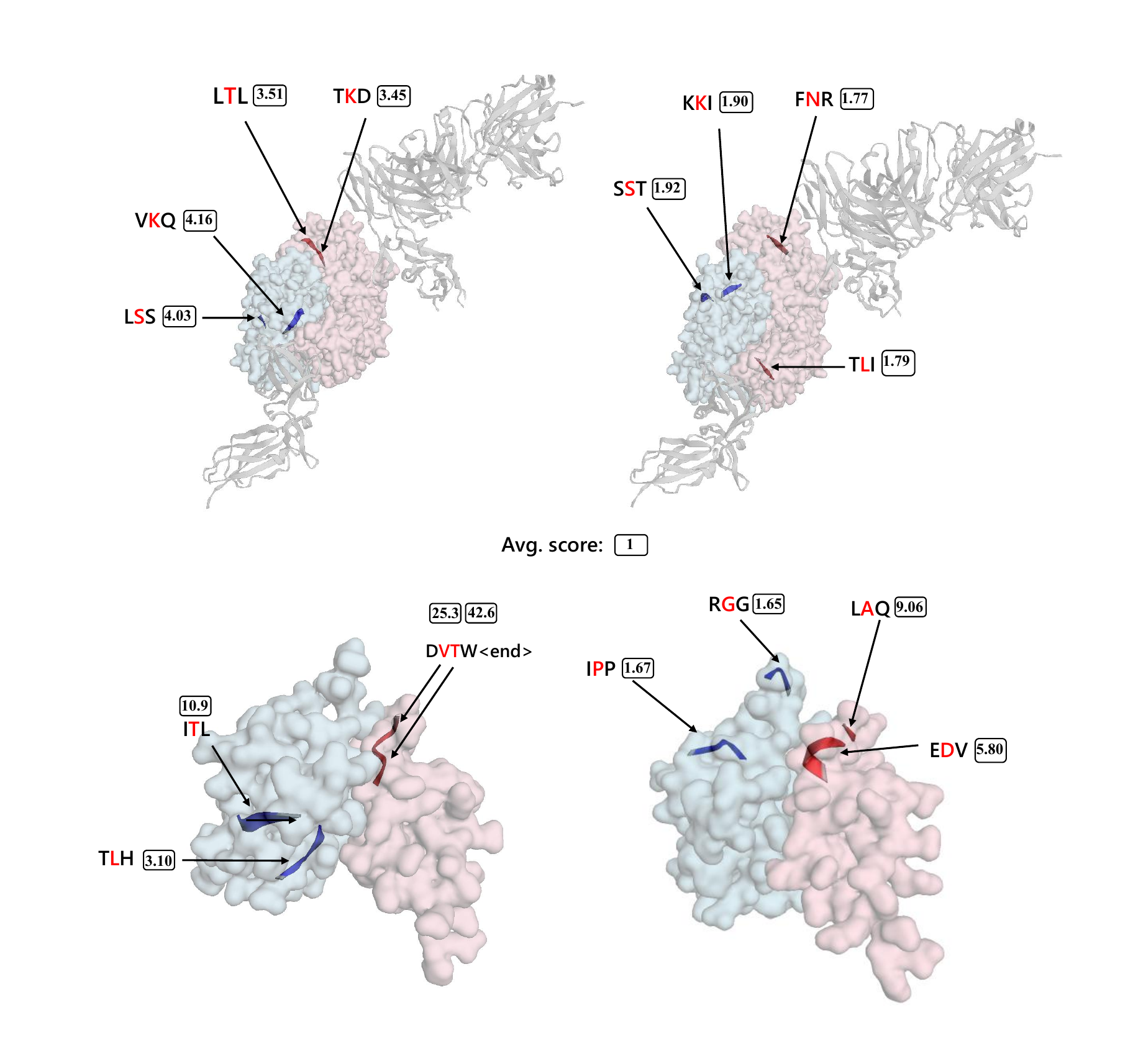}}
    \caption{The figure displays the protein-protein interaction complexes of the Human Tissue Factor (PDB ID: 1ahw) and Human Fanconi anemia-associated protein (PDB ID: 2MUR). Results from~\name~and Transformer are shown on the left and right, respectively. The top 2 scoring residues in the cross-attention matrix for Chain-A and Chain-B are colored in red and blue. The surfaces of Chain-A and Chain-B are highlighted in pink and light blue. The score represents the response scores of the top 2 residues.}
    \label{fig: Protein graph}
\end{figure}
To explore the alignment of protein pairs in their attention space, we randomly select a protein sequence from Subcellular localization Task, as well as a positive and negative protein associated with it. For protein-wise tasks, the positive protein is specifically chosen from proteins that share the same class. Figure~\ref{fig: attention matrix} displays cross-attention matrices for positive and negative protein pairs under our framework. In the matrices for positive pairs, larger attention scores at certain residues suggest their significant role in protein alignment within the same category.
Our method, in contrast to using a single-layer transformer classifier post-PLM encoding, produces cross-attention matrices that more clearly differentiate between positive and negative samples. 
As depicted in the figure, positive samples display elevated attention scores for specific residues, whereas negative samples demonstrate a relatively uniform distribution of attention across the matrix.  
This observation suggests that our approach forces a pair of negative protein sequences to fail to align in attention space which contributes to predicting the two proteins have different properties or functions. \\

\noindent Additionally, to visualize what patterns these attentions will discover, we conducted an experimental analysis by randomly selecting two pairs of \textbf{interactive protein sequences} from the test set of Human PPI. 
We employed a 3D folded structure visualization technique to display the protein interaction complex, as depicted in Figure~\ref{fig: Protein graph}. In this figure, the light blue and pink shades represent the interfaces through which different proteins interact within the complex. The intersection of these interfaces corresponds to the amino acid residue site where the two proteins bind. To identify the most responsive amino acid residue, we labeled the two amino acid residues within each protein structure that obtained the highest response scores when interacting with the other protein. These response scores were computed by summing the attention fractions of each amino acid residue across the entirety of the other protein. Since the attention matrix is computed by the softmax function, the amino acid residue response score will be 1. To enhance clarity, we also labeled two additional amino acid residues adjacent to each selected amino acid residues. Our observations revealed that the \name~outperformed traditional Transformers in terms of interpretability. \textbf{Notably, the~\name~assigned higher weights to amino acid residues in proximity to the protein-binding interface}, where the pink and light blue interfaces intersected. This behavior was not achieved by the traditional transformer, thus providing further validation of the efficacy of negative sample mining technology for PLM.

\section{Conclusion}
In this work, we introduced the NM-Transformer, a cheap but efficient approach that augments Protein Language Models (PLMs) by integrating negative sample mining to enhance functional sequence discrimination beyond evolutionary constraints. Our empirical results, validated across various protein-related tasks, confirm that our model significantly boosts both the performance and interpretability of existing PLMs, especially when trained from scratch. NM-Transformer enables PLMs to generate discriminative cross-attention matrices for positive and negative sample pair sequences. The cross-attention score exhibits the capability to identify amino acid residues near the binding interface, suggesting that our method aids PLMs in gaining deeper insights into the mechanism of protein interactions with practical application potential.  

\begin{thebibliography}{10}
\providecommand{\url}[1]{\texttt{#1}}
\providecommand{\urlprefix}{URL }
\providecommand{\doi}[1]{https://doi.org/#1}

\bibitem{alley2019unified}
Alley, E.C., Khimulya, G., Biswas, S., AlQuraishi, M., Church, G.M.: Unified rational protein engineering with sequence-based deep representation learning. Nature methods  \textbf{16}(12),  1315--1322 (2019)

\bibitem{baek2021accurate}
Baek, M., DiMaio, F., Anishchenko, I., Dauparas, J., Ovchinnikov, S., Lee, G.R., Wang, J., Cong, Q., Kinch, L.N., Schaeffer, R.D., et~al.: Accurate prediction of protein structures and interactions using a three-track neural network. Science  \textbf{373}(6557),  871--876 (2021)

\bibitem{bairoch2000enzyme}
Bairoch, A.: The enzyme database in 2000. Nucleic acids research  \textbf{28}(1),  304--305 (2000)

\bibitem{bepler2021learning}
Bepler, T., Berger, B.: Learning the protein language: Evolution, structure, and function. Cell systems  \textbf{12}(6),  654--669 (2021)

\bibitem{bonetta2020machine}
Bonetta, R., Valentino, G.: Machine learning techniques for protein function prediction. Proteins: Structure, Function, and Bioinformatics  \textbf{88}(3),  397--413 (2020)

\bibitem{burley2017protein}
Burley, S.K., Berman, H.M., Kleywegt, G.J., Markley, J.L., Nakamura, H., Velankar, S.: Protein data bank (pdb): the single global macromolecular structure archive. Protein crystallography: methods and protocols pp. 627--641 (2017)

\bibitem{chen2016profold}
Chen, D., Tian, X., Zhou, B., Gao, J., et~al.: Profold: Protein fold classification with additional structural features and a novel ensemble classifier. BioMed research international  \textbf{2016} (2016)

\bibitem{dallago2021flip}
Dallago, C., Mou, J., Johnston, K.E., Wittmann, B.J., Bhattacharya, N., Goldman, S., Madani, A., Yang, K.K.: Flip: Benchmark tasks in fitness landscape inference for proteins. Advances in Neural Information Processing Systems pp. 2021--11 (2021)

\bibitem{de2013emerging}
De~Juan, D., Pazos, F., Valencia, A.: Emerging methods in protein co-evolution. Nature Reviews Genetics  \textbf{14}(4),  249--261 (2013)

\bibitem{devlin2018bert}
Devlin, J., Chang, M.W., Lee, K., Toutanova, K.: Bert: Pre-training of deep bidirectional transformers for language understanding. arXiv preprint arXiv:1810.04805  (2018)

\bibitem{elnaggar2021prottrans}
Elnaggar, A., Heinzinger, M., Dallago, C., Rehawi, G., Wang, Y., Jones, L., Gibbs, T., Feher, T., Angerer, C., Steinegger, M., et~al.: Prottrans: Toward understanding the language of life through self-supervised learning. TPAMI  \textbf{44}(10) (2021)

\bibitem{ProtBert}
Elnaggar, A., Heinzinger, M., Dallago, C., Rehawi, G., Yu, W., Jones, L., Gibbs, T., Feher, T., Angerer, C., Steinegger, M., Bhowmik, D., Rost, B.: Prottrans: Towards cracking the language of lifes code through self-supervised deep learning and high performance computing. TPAMI pp.~1--1 (2021)

\bibitem{evans2021protein}
Evans, R., O’Neill, M., Pritzel, A., Antropova, N., Senior, A., Green, T., {\v{Z}}{\'\i}dek, A., Bates, R., Blackwell, S., Yim, J., et~al.: Protein complex prediction with alphafold-multimer. biorxiv pp. 2021--10 (2021)

\bibitem{fang2023method}
Fang, X., Wang, F., Liu, L., He, J., Lin, D., Xiang, Y., Zhu, K., Zhang, X., Wu, H., Li, H., et~al.: A method for multiple-sequence-alignment-free protein structure prediction using a protein language model. Nature Machine Intelligence  \textbf{5}(10),  1087--1096 (2023)

\bibitem{ferruz2022controllable}
Ferruz, N., H{\"o}cker, B.: Controllable protein design with language models. Nature Machine Intelligence  \textbf{4}(6),  521--532 (2022)

\bibitem{gligorijevic2021structure}
Gligorijevi{\'c}, V., Renfrew, P.D., Kosciolek, T., Leman, J.K., Berenberg, D., Vatanen, T., Chandler, C., Taylor, B.C., Fisk, I.M., Vlamakis, H., et~al.: Structure-based protein function prediction using graph convolutional networks. Nature communications  \textbf{12}(1), ~3168 (2021)

\bibitem{heinzinger2019modeling}
Heinzinger, M., Elnaggar, A., Wang, Y., Dallago, C., Nechaev, D., Matthes, F., Rost, B.: Modeling aspects of the language of life through transfer-learning protein sequences. BMC bioinformatics  \textbf{20}(1),  1--17 (2019)

\bibitem{hie2021learning}
Hie, B., Zhong, E.D., Berger, B., Bryson, B.: Learning the language of viral evolution and escape. Science  \textbf{371}(6526),  284--288 (2021)

\bibitem{hie2022evolutionary}
Hie, B.L., Yang, K.K., Kim, P.S.: Evolutionary velocity with protein language models predicts evolutionary dynamics of diverse proteins. Cell Systems  \textbf{13}(4),  274--285 (2022)

\bibitem{hsu2022learning}
Hsu, C., Verkuil, R., Liu, J., Lin, Z., Hie, B., Sercu, T., Lerer, A., Rives, A.: Learning inverse folding from millions of predicted structures. In: ICML. pp. 8946--8970. PMLR (2022)

\bibitem{hu2022exploring}
Hu, M., Yuan, F., Yang, K., Ju, F., Su, J., Wang, H., Yang, F., Ding, Q.: Exploring evolution-aware \&-free protein language models as protein function predictors. Advances in Neural Information Processing Systems  \textbf{35},  38873--38884 (2022)

\bibitem{jones2012psicov}
Jones, D.T., Buchan, D.W., Cozzetto, D., Pontil, M.: Psicov: precise structural contact prediction using sparse inverse covariance estimation on large multiple sequence alignments. Bioinformatics  \textbf{28}(2),  184--190 (2012)

\bibitem{jumper2021highly}
Jumper, J., Evans, R., Pritzel, A., Green, T., Figurnov, M., Ronneberger, O., Tunyasuvunakool, K., Bates, R., {\v{Z}}{\'\i}dek, A., Potapenko, A., et~al.: Highly accurate protein structure prediction with alphafold. Nature  \textbf{596}(7873),  583--589 (2021)

\bibitem{khurana2018deepsol}
Khurana, S., Rawi, R., Kunji, K., Chuang, G.Y., Bensmail, H., Mall, R.: Deepsol: a deep learning framework for sequence-based protein solubility prediction. Bioinformatics  \textbf{34}(15),  2605--2613 (2018)

\bibitem{lin2022language}
Lin, Z., Akin, H., Rao, R., Hie, B., Zhu, Z., Lu, W., dos Santos~Costa, A., Fazel-Zarandi, M., Sercu, T., Candido, S., et~al.: Language models of protein sequences at the scale of evolution enable accurate structure prediction. Science  (2023)

\bibitem{lin2023evolutionary}
Lin, Z., Akin, H., Rao, R., Hie, B., Zhu, Z., Lu, W., Smetanin, N., Verkuil, R., Kabeli, O., Shmueli, Y., et~al.: Evolutionary-scale prediction of atomic-level protein structure with a language model. Science  \textbf{379}(6637),  1123--1130 (2023)

\bibitem{marks2011protein}
Marks, D.S., Colwell, L.J., Sheridan, R., Hopf, T.A., Pagnani, A., Zecchina, R., Sander, C.: Protein 3d structure computed from evolutionary sequence variation. PloS one  \textbf{6}(12),  e28766 (2011)

\bibitem{meng2023improved}
Meng, Q., Guo, F., Tang, J.: Improved structure-related prediction for insufficient homologous proteins using msa enhancement and pre-trained language model. Briefings in Bioinformatics  \textbf{24}(4),  bbad217 (2023)

\bibitem{rajagopal2003subcellular}
Rajagopal, A., Simon, S.M.: Subcellular localization and activity of multidrug resistance proteins. Molecular biology of the cell  \textbf{14}(8),  3389--3399 (2003)

\bibitem{rao2016noise}
Rao, J., He, H., Lin, J.: Noise-contrastive estimation for answer selection with deep neural networks. In: Proceedings of the 25th ACM International on Conference on Information and Knowledge Management. pp. 1913--1916 (2016)

\bibitem{rao2019evaluating}
Rao, R., Bhattacharya, N., Thomas, N., Duan, Y., Chen, P., Canny, J., Abbeel, P., Song, Y.: Evaluating protein transfer learning with tape. Advances in neural information processing systems  \textbf{32} (2019)

\bibitem{rao2020transformer}
Rao, R., Meier, J., Sercu, T., Ovchinnikov, S., Rives, A.: Transformer protein language models are unsupervised structure learners. Biorxiv pp. 2020--12 (2020)

\bibitem{rao2021msa}
Rao, R.M., Liu, J., Verkuil, R., Meier, J., Canny, J., Abbeel, P., Sercu, T., Rives, A.: Msa transformer. In: ICML. pp. 8844--8856. PMLR (2021)

\bibitem{rives2021biological}
Rives, A., Meier, J., Sercu, T., Goyal, S., Lin, Z., Liu, J., Guo, D., Ott, M., Zitnick, C.L., Ma, J., et~al.: Biological structure and function emerge from scaling unsupervised learning to 250 million protein sequences. Proceedings of the National Academy of Sciences  \textbf{118}(15),  e2016239118 (2021)

\bibitem{shrivastava2016training}
Shrivastava, A., Gupta, A., Girshick, R.: Training region-based object detectors with online hard example mining. In: Proceedings of the IEEE conference on computer vision and pattern recognition. pp. 761--769 (2016)

\bibitem{strokach2020fast}
Strokach, A., Becerra, D., Corbi-Verge, C., Perez-Riba, A., Kim, P.M.: Fast and flexible protein design using deep graph neural networks. Cell systems  \textbf{11}(4) (2020)

\bibitem{wang2023scientific}
Wang, H., Fu, T., Du, Y., Gao, W., Huang, K., Liu, Z., Chandak, P., Liu, S., Van~Katwyk, P., Deac, A., et~al.: Scientific discovery in the age of artificial intelligence. Nature  \textbf{620}(7972),  47--60 (2023)

\bibitem{wang2020reinforced}
Wang, X., Xu, Y., He, X., Cao, Y., Wang, M., Chua, T.S.: Reinforced negative sampling over knowledge graph for recommendation. In: WWW. pp. 99--109 (2020)

\bibitem{wang2024hierarchical}
Wang, Y., Song, J., Dai, Q., Duan, X.: Hierarchical negative sampling based graph contrastive learning approach for drug-disease association prediction. IEEE Journal of Biomedical and Health Informatics  (2024)

\bibitem{wei2017improved}
Wei, L., Xing, P., Zeng, J., Chen, J., Su, R., Guo, F.: Improved prediction of protein--protein interactions using novel negative samples, features, and an ensemble classifier. Artificial Intelligence in Medicine  \textbf{83},  67--74 (2017)

\bibitem{xu2023protst}
Xu, M., Yuan, X., Miret, S., Tang, J.: Protst: Multi-modality learning of protein sequences and biomedical texts. ICML  (2023)

\bibitem{xu2022peer}
Xu, M., Zhang, Z., Lu, J., Zhu, Z., Zhang, Y., Ma, C., Liu, R., Tang, J.: Peer: A comprehensive and multi-task benchmark for protein sequence understanding. NIPS  (2022)

\bibitem{ying2018graph}
Ying, R., He, R., Chen, K., Eksombatchai, P., Hamilton, W.L., Leskovec, J.: Graph convolutional neural networks for web-scale recommender systems. In: SIGKDD. pp. 974--983 (2018)

\bibitem{yu2023enzyme}
Yu, T., Cui, H., Li, J.C., Luo, Y., Jiang, G., Zhao, H.: Enzyme function prediction using contrastive learning. Science  (2023)

\bibitem{zhang2024protein}
Zhang, Z., Wayment-Steele, H.K., Brixi, G., Wang, H., Dal~Peraro, M., Kern, D., Ovchinnikov, S.: Protein language models learn evolutionary statistics of interacting sequence motifs. bioRxiv pp. 2024--01 (2024)

\bibitem{zhang2022protein}
Zhang, Z., Xu, M., Jamasb, A., Chenthamarakshan, V., Lozano, A., Das, P., Tang, J.: Protein representation learning by geometric structure pretraining. In: ICLR (2023)

\bibitem{zheng2023structure}
Zheng, Z., Deng, Y., Xue, D., Zhou, Y., Ye, F., Gu, Q.: Structure-informed language models are protein designers. ICML pp. 2023--02 (2023)

\end{thebibliography}
\end{document}